\documentclass{article}
\usepackage{amssymb}
\usepackage{amsmath, amssymb}
\usepackage{graphicx}
\usepackage{booktabs}
\usepackage{enumitem} 
\usepackage{wrapfig} 
\usepackage{array}
\usepackage{multirow}
\usepackage{longtable}
\usepackage{needspace}
\usepackage{ulem}

\usepackage[dvipsnames]{xcolor}
\usepackage[preprint]{corl_2025} 

\newcommand{\methodname}[1]{\textsc{Scizor}}

\title{\methodname{}: A Self-Supervised Approach to Data Curation for Large-Scale Imitation Learning}

\author{
Yu Zhang$^{1*}$, 
Yuqi Xie$^{1,2*}$, 
Huihan Liu$^{1\dagger}$, 
Rutav Shah$^{1\dagger}$, \\[5pt] 
\textbf{Michael Wan$^{\ddagger2}$, 
Linxi ``Jim" Fan$^2$, 
Yuke Zhu$^{1,2}$} \\[5pt]
$^1$The University of Texas at Austin ~ $^2$NVIDIA Research
}

\begin{document}
\maketitle

\let\thefootnote\relax
\footnotetext[0]{\makebox[0pt][r]{\textsuperscript{*, $\dagger$} }Equal contribution.}
\footnotetext[0]{\makebox[0pt][r]{\textsuperscript{$\ddagger$} }This work was done while Michael Wan was interning at NVIDIA.}



\vspace{-2mm}
\begin{abstract}
Imitation learning advances robot capabilities by enabling the acquisition of diverse behaviors from human demonstrations. However, large-scale datasets used for policy training often introduce substantial variability in quality, which can negatively impact performance. As a result, automatically curating datasets by filtering low-quality samples to improve quality becomes essential. Existing robotic curation approaches rely on costly manual annotations and perform curation at a coarse granularity, such as the dataset or trajectory level, failing to account for the quality of individual state-action pairs. To address this, we introduce \methodname{}, 
the first self-supervised transition-level curation framework that requires no annotations and scales to large-scale datasets to improve the performance of imitation learning policies and modern Vision-Language-Action (VLA) models. \methodname{} targets two complementary sources of low-quality data: \textit{suboptimal} data, which hinders learning with undesirable actions, and \textit{redundant} data, which dilutes training with repetitive patterns. \methodname{} leverages a self-supervised task progress predictor for suboptimal data to remove samples lacking task progression, and a deduplication module operating on joint state-action representation for samples with redundant patterns. Empirically, we show that \methodname{} enables imitation learning policies and modern VLA models to achieve higher performance with less data, yielding an average improvement of $15.4$\% across multiple benchmarks. More information is available at: \url{https://ut-austin-rpl.github.io/SCIZOR/}
\end{abstract}


\section{Introduction}
Imitation learning has shown promising signs in acquiring a wide range of motor behaviors by learning from expert demonstrations, a necessary step towards general-purpose robots. Recent advances in Vision-Language-Action (VLA) models have highlighted the potential of large-scale imitation learning, where massive datasets spanning diverse tasks and environments are used to train generalist policies. 
Such diverse, large-scale data collection inherently introduces variability in data quality \citep{mandlekar2021matterslearningofflinehuman, brown2019betterthandemonstratorimitationlearningautomaticallyranked, lin2025datascalinglawsimitation}, including mistakes made by operators leading to suboptimal actions (\textit{e.g.}, dropping an object), or redundancy in data leading to skewed distributions. Such a dataset can misguide models into learning incorrect behaviors \citep{mandlekar2021matterslearningofflinehuman,liu2022robot} and hinder diversity \citep{lin2025datascalinglawsimitation}, reducing the impact of rare but informative actions. Therefore, effective data curation, the process of filtering data to improve the data quality \citep{hejna2024re, hejna2025robotdatacurationmutual}, becomes critical for building robust and high-performing imitation learning policies.

\begin{figure}[t]
    \centering
    \includegraphics[width=.99\linewidth]{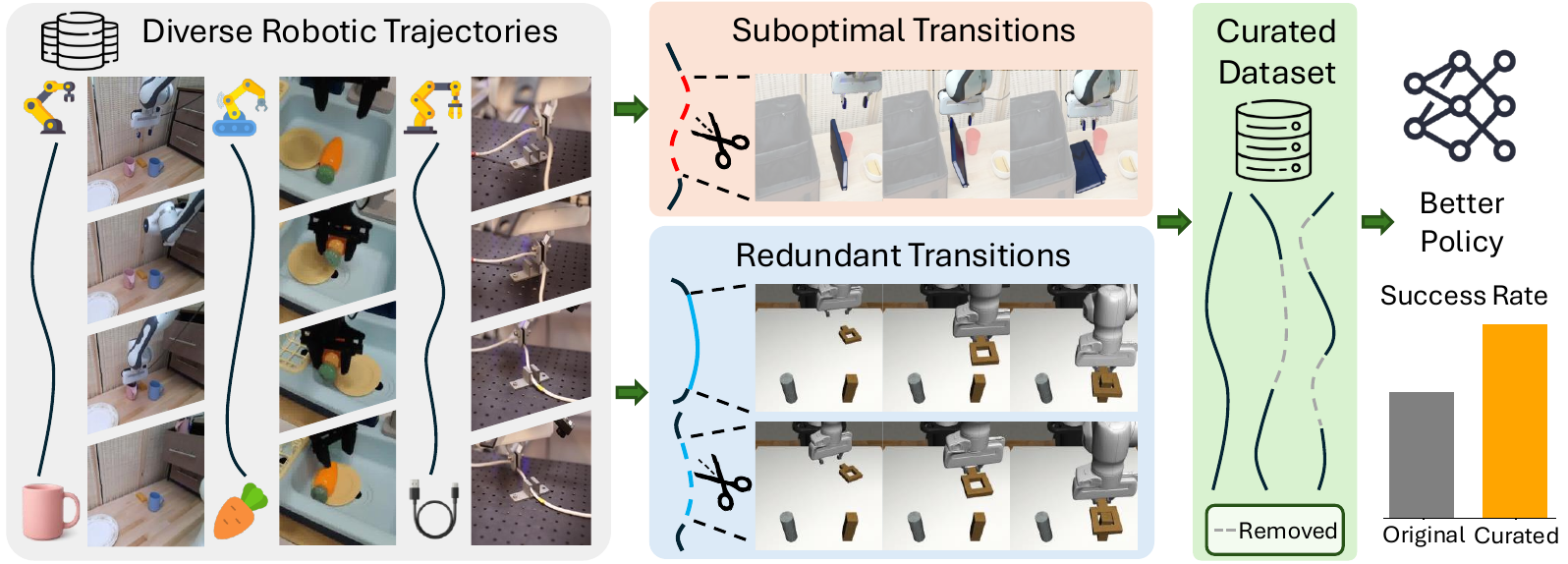}
    \caption{\textbf{\methodname{} overview.} Each trajectory from the original robotic datasets is simultaneously passed through the suboptimal transitions removal module and the redundant transitions removal module. Each module removes data
    , resulting in a curated dataset. A policy trained on the curated dataset achieves a higher success rate.}

    \label{fig:method_figure}
    \vspace{-5mm}
\end{figure}


Early efforts in robotic data curation have relied heavily on human annotations to label high- and low-quality data \citep{mandlekar2021matterslearningofflinehuman}, but these methods have largely been confined to small-scale datasets. As data scales up, manual annotation becomes infeasible, making it important to \textit{automatically} curate data, an approach that has already shown promise in fields like computer vision (CV) and natural language processing (NLP) \citep{abbas2023semdedup, albalak2024surveydataselectionlanguage, Schuhmann2022LAION5BAO, Xie2023DoReMiOD}. Specifically in robot learning, a key challenge in this process to maximize data utilization is \textit{the need for curating at the finest granularity: we must evaluate the quality of individual state-action pairs}. For instance, a trajectory may include an initial failed grasp attempt followed by a successful recovery, containing both suboptimal and valuable segments. Effective curation should isolate and remove only the uninformative or erroneous segments, like the failed grasp, while preserving segments that provide useful learning signals. 
The different impact of individual data points for learning has also been underscored in prior work, such as weighted behavior cloning \citep{liu2022robot, wang2020critic, Kostrikov2021OfflineRL}.
However, current large-scale imitation learning curation methods have yet to address data quality at a fine-grained level. Existing approaches typically curate by reweighting entire dataset domains \citep{octomodelteam2024octoopensourcegeneralistrobot, hejna2024re} or discarding entire trajectories \citep{hejna2025robotdatacurationmutual, chen2025curatingdemonstrationsusingonline}, overlooking the contribution and quality of individual state-action pairs. 

Effectively curating individual state-action pairs is challenging, especially in large-scale datasets, as robot demonstrations typically lack dense reward annotations, making it difficult to assess the quality of every interaction step. To address this, we develop a \textit{self-supervised} approach for filtering low-quality state-action pairs, offering a scalable solution for improving data quality in imitation learning.
Our work is motivated by two key observations: (1) \textit{suboptimal} transitions, which contain undesirable actions like collision, jittering, and other erroneous actions,
can degrade policy performance by reinforcing incorrect behaviors; and (2) \textit{redundant} transitions, which repeat common patterns excessively, can dilute the learning signal by dominating other informative and diverse samples.


We introduce \methodname{}, the first  \textbf{self-supervised, transition-level data curation method that scales to the Open-X Embodiment Dataset~\cite{padalkar2023open} with more than one million trajectories for large-scale models such as the Octo~\cite{octomodelteam2024octoopensourcegeneralistrobot} vision-language-action (VLA) model of over 27M parameters}.
\methodname{} is a \textbf{\underline{s}}elf-supervised data \textbf{\underline{c}}uration method that reduces dataset s\textbf{\underline{iz}}e by filtering out sub\textbf{\underline{o}}ptimal and \textbf{\underline{r}}edundant state-action pairs.
First, to identify suboptimal data without access to reward information, we train a self-supervised task progress predictor using temporal distance classification \citep{aytar2018playing, sermanet2017unsupervisedperceptualrewardsimitation, zakka2021xirlcrossembodimentinversereinforcement}, and remove frames that do not demonstrate meaningful progress toward the task goal. For instance, imagine a robot attempting to grasp an object and failing, then immediately trying again and succeeding. The initial failed attempt produces little or even negative progress, so it is discarded. The subsequent successful grasp, however, still advances the task and is retained.
Second, to remove redundant data, a key insight is that some segments may appear visually similar while differing substantially in the executed actions, and vice versa.
Therefore, both visual observations and their corresponding actions must be considered for effective deduplication. To this end, we apply deduplication \citep{abbas2023semdedup} using joint representations of state and action to identify and filter redundant state-action pairs.
We then filter out frames based on similarity scores to reduce repetition while preserving dataset diversity.
In summary, the suboptimal frame filter targets harmful or noisy supervision, while the redundancy filter removes overrepresented patterns. Together, the two deletion strategies complement each other by targeting distinct modes of low-quality data. 

In summary, our key contributions are as follows: 
\vspace{-2pt}
\begin{itemize}[itemsep=0pt, topsep=0pt, parsep=1pt, partopsep=0pt, left=0pt]
\item We propose a unified framework for transition-level data curation that filters both suboptimal and redundant state-action pairs in large-scale imitation learning datasets for imitation learning policies and modern Vision-Language-Action models. 
\item We introduce a suboptimality detector based on self-supervised task progress estimation, and a deduplication module that removes repetitive data to preserve data diversity. 
\item We empirically demonstrate that \methodname{} improves policy performance across diverse large-scale imitation learning benchmarks, showing on average 15.4\% improvement.
\end{itemize}
\vspace{-5pt}

\vspace{-6pt}
\section{Related Work}
\vspace{-6pt}

\paragraph{Imitation Learning on Large-Scale Robot Datasets.}
Imitation learning has been a popular approach to learn robot policy from human demonstrations \citep{pomerleau1989alvinn,zhang2017deep,mandlekar2021matterslearningofflinehuman,florence2021implicit} to scale up robot policy generalization and enable diverse behaviors, there has recently been progress in large-scale multi-task imitation learning \citep{brohan2023rt1roboticstransformerrealworld, brohan2023rt, octomodelteam2024octoopensourcegeneralistrobot, kim24openvla, black2024pi0visionlanguageactionflowmodel, nvidia2025gr00tn1openfoundation} trained on robot trajectory data of a wide variety of tasks. This progress is driven not only by advances in policy architectures \citep{brohan2023rt1roboticstransformerrealworld, mandlekar2021matterslearningofflinehuman, ho2020denoising, chi2024diffusion}, but more importantly the collection of large-scale datasets in both real-world \citep{ebert2021bridge, padalkar2023open, khazatsky2024droid, contributors2025agibotworld} and simulation \citep{mandlekar2023mimicgen, nasiriany2024robocasa}. 
These datasets are often collected from multiple institutions using varied hardware configurations and teleoperation systems \citep{walke2024bridgedatav2datasetrobot, aloha2team2024aloha, iyer2024open, fu2024mobile, zhao2023learning, lin2024learning, zhao2023learning}, resulting in inconsistencies in quality and redundancy across different datasets. Although robotics datasets have been scaled to unprecedented sizes, the study of dataset quality and data curation methods remains preliminary.
\vspace{-10pt}


 
\paragraph{Data Curation in Vision and Language Models.}
Data curation, which is the selection and filtering of data for better training results, have been extensively studied in both computer vision and language modeling to address the challenges posed by large-scale, heterogeneous datasets \citep{albalak2024surveydataselectionlanguage}. In vision, LAION-5B \citep{Schuhmann2022LAION5BAO} uses pretrained encoders like CLIP to assign data quality on the samples. In language modeling, data mixture strategies like DoReMi \citep{Xie2023DoReMiOD} balance various data sources for distribution robustness, while deduplication methods like SemDeDup \citep{abbas2023semdedup} remove near-duplicates using semantic embeddings. Data Filtering Networks \citep{fang2023datafilteringnetworks} trains a neural network to distinguish informative versus less-informative data, while Ask-LLM \citep{sachdeva2024traindataefficientllms} uses instruction-tuned LLMs to assess the quality of training examples directly. Meanwhile, Less-Is-More-RL \citep{li2025limrrlscaling} shows how pruning suboptimal data can improve downstream policy performance in reinforcement learning settings. 
\vspace{-10pt}







\paragraph{Data Curation for Robotics.} Data quality has been known to affect policy learning performance for robotics \citep{mandlekar2021matterslearningofflinehuman, belkhale2023dataqualityimitationlearning}. 
There have been studies in improving human demonstration quality, albeit in small-scale tasks, by automatic ranking \citep{brown2019betterthandemonstratorimitationlearningautomaticallyranked}, or eliciting compatible behavior from humans during the data collection process \citep{Gandhi2022ElicitingCD}. 
As progress in general-purpose, large-scale robot learning continues, there has been growing interest in curating large-scale datasets for robot learning \citep{octomodelteam2024octoopensourcegeneralistrobot, kim24openvla, hejna2024re, hejna2025robotdatacurationmutual, chen2025curatingdemonstrationsusingonline}. Octo \citep{octomodelteam2024octoopensourcegeneralistrobot} and OpenVLA \citep{kim24openvla} perform ad-hoc \textit{dataset-level} curation by heuristically tuning a set of weights for data mixtures, balancing the dataset composition; Remix \citep{hejna2024re} automates this dataset-level curation with distributionally robust optimization. DemInf \citep{hejna2025robotdatacurationmutual} performs \textit{trajectory-level} curation with mutual information as a trajectory quality estimator, and Demo-SCORE \citep{chen2025curatingdemonstrationsusingonline} also performs trajectory-level curation, but has to rely on online rollout performance. DataMIL~\citep{dass2025datamilselectingdatarobot} select task relevant trajectories from Open-X~\cite{padalkar2023open} datasets to boost performance on task-specific policies.

\vspace{-5pt}



\vspace{-6pt}
\section{Self-Supervised Data Curation for Large-Scale Imitation Learning}
\vspace{-6pt}
\begin{figure}[t]
    \centering
    \includegraphics[width=.99\linewidth]{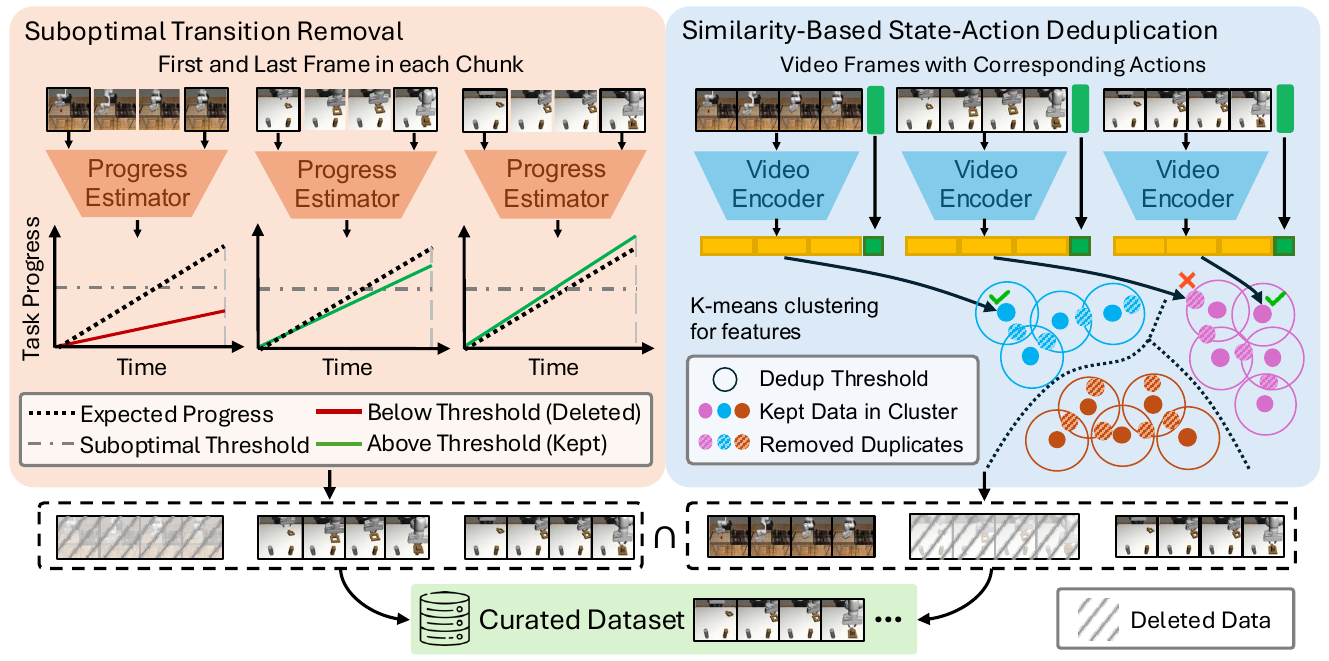}
    \caption{
    \textbf{\methodname{}'s architecture.} 
    We apply two curation modules: 
    (1) \emph{Suboptimal transition removal}, where we estimate chunk progress from its first and last frames and discard those below a threshold; 
    (2) \emph{State-action deduplication}, where we encode all frames, cluster their features via K-means, and remove frames whose intra-cluster cosine similarity exceeds a threshold.}
    \label{fig:method_figure}
    \vspace{-10pt}
\end{figure}
\vspace{-2pt}
We introduce our data curation framework, \methodname{}, which performs fine-grained filtering of low-quality data in a self-supervised manner to improve imitation learning policy performance. We begin by introducing key formulations and background, followed by two core components of our method: (1) a self-supervised suboptimal transitions removal module; (2) a similarity-based state-action deduplication module that filters redundant transitions.
\vspace{-5pt}
\subsection{Preliminaries and Formulations}
\vspace{-5pt}
We formulate a robot manipulation task as a Markov Decision Process $\mathcal{M} = (\mathcal{S}, \mathcal{A}, \mathcal{R}, \mathcal{P}, p_{0}, \gamma )$ representing the state space, action space, reward function, transition probability, initial state distribution, and discount factor. Given the current state $s_t\in\mathcal{S}$, the robot action $a_t\in\mathcal{A}$ is drawn from the policy $\pi\left(\cdot \mid s_{t}\right)$. 
The objective of imitation learning is to learn a policy $\pi_{R}$ parameterized by $\theta$ that maximizes the log-likelihood of actions $a$ conditioned on the states $s$:
\vspace{-5pt}
\begin{equation}
\theta^*=\underset{\theta}{\arg \max}  \underset{(s, a) \sim \mathcal{D}_{expert}}{\mathbb{E}}\left[\log \pi_{\theta}(a \mid s)\right]
\end{equation}
where $(s,a)$ are samples from the human demonstration dataset $\mathcal{D}_{expert}$.
Our data curation objective is to refine $\mathcal{D}_{expert}$ by filtering out suboptimal or redundant samples to improve policy performance. 

\vspace{-5pt}
\subsection{Suboptimal Transition Removal via Progress Estimation}
\vspace{-5pt}



\label{sec:method-subop}
    Human demonstrations often contain both proficient and suboptimal segments in the same trajectory with no explicit signals. Manually labeling suboptimal segments is labor-intensive and not scalable. We propose a self-supervised approach to detect suboptimal behaviors based on the intuition that \textit{progress toward task completion should increase steadily over time}. 
    As we don’t have ground-truth progress labels, we use the ground truth time elapsed between two observations as a stand-in for the task progress. We train a lightweight model to predict this progress from pairs of states. 
    At test time, if the predicted progress for a segment is unexpectedly lower than expected progress, it can serve as a signal of suboptimality. This allows us to automatically identify and filter out segments that deviate from making progress, without requiring any manual annotations.

\textbf{Defining Suboptimality with Task Progress.} Inspired by temporal distance classification in self-supervised representation learning \citep{aytar2018playing}, we evaluate action quality by estimating the task progress between two timesteps.
If the predicted progress is significantly lower than the elapsed time, this means the robot is behind schedule (i.e., progressing less than expected), meaning that the sub-trajectory is suboptimal.
Specifically, we define a progress function $f:S_{i:i+T}\rightarrow T_p$, which inputs a sub-trajectory $S_{i:i+T}$ from timestep $i$ to $i+T$, and predicts the progress $T_p$ made over the sub-trajectory $S_{i:i+T}$ towards completion. \textit{$T_p$ measures the temporal distance that the robot has moved the task forward over the sub‑trajectory $S_{i:i+T}$, measured in seconds.}
We then compare this predicted progress $T_p$ to the actual elapsed time $T$. The suboptimality score for the sub-trajectory $S_{i:i+T}$ is defined as $V_{i:i+T}=T-T_p$.
\textbf{Assigning Suboptimal Scores to Individual Samples.}
Our goal is to assign a suboptimality score to each individual transition, enabling fine-grained data filtering. 
While our scores are initially predicted at the sub-trajectory level, we want each transition’s score to include 3 factors: 
(1) Progress predicted from each sub-trajectory that current transition belongs to, 
(2) Future suboptimality with reducing impact over time, 
so that the current transition score accounts for suboptimal scenarios it leads to, 
(3) The overall quality of the entire trajectory that the transition belongs to. 
This enables a transition’s score to reflect both its local impact and the global quality of the whole trajectory.

Formally, from the previous section we have sub-trajectory scores $V_{0:T}, V_{1:1+T}, \dots, V_{N:N+T}$. 
Assuming equal contribution from each transition in a sub-trajectory, a score $V_{i:i+T}$ is evenly distributed across its $T$ constituent transitions, contributing $\frac{1}{T} V_{i:i+T}$ to each. 
The aggregated sample-level score is computed as
$
\hat{V_i} = \sum_{t=i-T}^{i} \frac{1}{T} V_{t:t+T}.
$
To allow the current suboptimality score to also account for future suboptimality with diminishing impact over time, we apply temporal discounting:
$
V_i = \sum_{t=i}^{T} \gamma^{t-i} \hat{V_t}, \quad \gamma \in [0, 1],
$
where $\gamma$ controls the rate at which future influences diminish. 
Finally, to incorporate trajectory-level quality, we combine each transition’s score $V_i$ with the mean score across all $N$ transitions in its trajectory, 
yielding the final curation score:
\vspace{-5pt}
\[
V_i^{\text{final}} = \alpha \cdot V_i + (1 - \alpha) \cdot \frac{1}{N} \sum_{j=1}^N V_j,
\]
where $\alpha \in [0,1]$ balances local and trajectory-wide quality.

\textbf{Removing Suboptimal Samples.} During policy training, we compute the suboptimality score for every sample as described above. During data curation, we exclude transitions with suboptimality scores above a certain threshold $\epsilon_s$ from the training process. 


\vspace{-8pt}
\subsection{Similarity-Based State-Action Deduplication}
\vspace{-5pt}

Large-scale imitation learning datasets often include many visually and behaviorally similar sequences, for example, repeated demonstrations of the same skill in nearly identical contexts. Training directly on all such data can hinder policy generalization by overemphasizing common patterns while underrepresenting rare but informative cases. To mitigate this, we introduce a similarity-based deduplication method that filters out redundant data.

A key insight is that some segments may appear visually similar, yet differ in task intent or executed actions. To avoid discarding meaningful variations, effective deduplication must consider both visual states and actions. To this end, we propose a similarity-based deduplication method that utilizes \textit{joint} representations of visual states and actions to identify and filter redundant state-action pairs.


\textbf{Defining State-Action Duplicates.}
Prior work on semantic deduplication \citep{abbas2023semdedup} has focused on curating large image datasets by removing semantically similar data pairs based solely on visual features. However, such visual-only deduplication methods are not well-suited for sequential decision-making tasks like imitation learning in robotics, where action dynamics play a crucial role. In this work, we extend the idea of semantic deduplication to the imitation learning domain by incorporating both visual states and action information. Specifically, we define state-action duplicates as state-action chunks $(S_{i:i+T}, a_{i:i+T})$ that are visually similar and lead to comparable actions, reflecting redundant patterns that contribute little to learning diversity.


\textbf{Generating State-Action Features.} We first divide the dataset into non-overlapping sub-trajectories, each consisting of a state-action sequence $(S_{i:i+T}, a_{i:i+T})$, where each chunk spans a fixed duration $T$. Given the variations in recording frequency across datasets, we uniformly subsample $N$ RGB images from each chunk for consistency. As raw visual data is high-dimensional and not directly suitable for similarity computation, we employ the Cosmos video encoder \citep{nvidia2025cosmosworldfoundationmodel}, a pre-trained model that encodes both temporal and semantic information from videos, to extract a compact 1D video feature vector $z_v$. We then concatenate the actions represented as delta end-effector pose to the visual embedding to form a joint state-action feature $z_{v+a}$.


\textbf{Removing Duplicated Samples.} 
We begin by performing K-means clustering to group semantically similar state-action chunks. Within each cluster, we compute pairwise cosine distances among all chunks. For each chunk, its similarity score is defined as the minimum distance it has with any other chunk in the same cluster.
We identify as duplicates those chunks whose maximum similarity exceeds a defined threshold, $\epsilon_{d}$, as they are highly similar to at least one other sample in their cluster. These chunks will be filtered out during policy training with a duplication mask.


\vspace{-8pt}
\section{Experiments}
\vspace{-8pt}


In our experiments, we aim to address the following questions: 
1) How much does \methodname{} improve imitation learning policy performance? 2) What advantage does \methodname{}'s fine-grained state-action curation offer over trajectory- or dataset-level curation in prior work? 3) What design components contribute most to \methodname{}? 4) What types of low-quality samples can \methodname{} identify and remove?
\vspace{-12pt}
\subsection{Experimental Setup}
\vspace{-8pt}

\textbf{Datasets and Training Details.}
We evaluate our method on three robotic benchmarks for imitation learning, chosen to represent a range of real-world scenarios: a large-scale crowdsourced dataset, a dataset featuring varying levels of human expertise, and a human-in-the-loop dataset with mixed data distributions. This selection enables us to evaluate \methodname{}'s effectiveness across various scenarios and diverse data regimes. For full dataset details, see Appendix~\ref{sec:dataset_detail}.
\vspace{-5pt}
\begin{itemize}[left=0pt, topsep=0pt, parsep=1pt, partopsep=0pt]
    \item \textbf{Open-X-Embodiment (OXE) \citep{padalkar2023open}}: A large-scale collection of over one million real-world robotic trajectories. We use the Simpler environment \citep{li24simpler} and benchmark on two tasks: \textit{Pick Can} and \textit{Move Near}. We train the Octo model \cite{octomodelteam2024octoopensourcegeneralistrobot} with two random seeds and use the same ``Magic Soup" weighting. This setting evaluates \methodname{}'s scalability to large and diverse datasets.
    \item \textbf{RoboMimic \citep{mandlekar2021matterslearningofflinehuman}}: A dataset and benchmark containing human-collected trajectories of varying proficiency. We use the simulated Multi-Human dataset for the \textit{Can} and \textit{Square} tasks to be consistent with the baseline comparison. We train the BC policy provided in the benchmark with three random seeds. This setting evaluates \methodname{}'s ability to curate demonstrations of mixed quality.
    \item \textbf{Sirius-Fleet \citep{liu2024multitaskinteractiverobotfleet}}: A real-world multi-task dataset comprising 1,500 policy rollouts with human interventions. Our real-world evaluation spans four task sets comprising eight tasks. We train the BC-Transformer policy used in the paper with three random seeds. This setting evaluates \methodname{}'s ability to curate mixed data from both autonomous policies and human corrections.
\end{itemize}

\textbf{Baselines.}
We benchmark \methodname{} against $3$ baselines, each highlighting a different aspect of data curation. We compare with \textbf{Uniform} to show the effectiveness of \methodname{}, with \textbf{DemoInf} and \textbf{Re-Mix} to show that fine-grained curation offers advantages over coarser filtering strategies.
\vspace{-5pt}

\begin{itemize}[left=0pt]
    \item \textbf{Uniform}: A baseline that uniformly deletes the same percentage of data as other methods to control for dataset size. This comparison ensures that the improvement observed with \methodname{} is attributed to which specific samples are removed, not simply to the reduced dataset size itself.
    \item \textbf{DemoInf \cite{hejna2025robotdatacurationmutual}}: A \textit{trajectory-level} method that estimates mutual information between states and actions for each trajectory as a quality score and removes low-quality trajectories.
    \item \textbf{Re-Mix \cite{hejna2024re}}: 
A \textit{dataset-level} method that learns data mixture weights for the ``RT-X" variant of the OXE datasets. To ensure consistency, we train the Octo-small model on OXE\textsubscript{RT-X} for \methodname{}, while directly adopting their learned weights for Re-Mix.

\end{itemize}
\vspace{-5pt}



\vspace{-8pt}
\subsection{Experimental Results}
\vspace{-8pt}

\textbf{RQ1: How much does \methodname{} improve imitation learning policy performance?}
\begin{figure}[t]
    \centering
    \includegraphics[width=.98\linewidth]{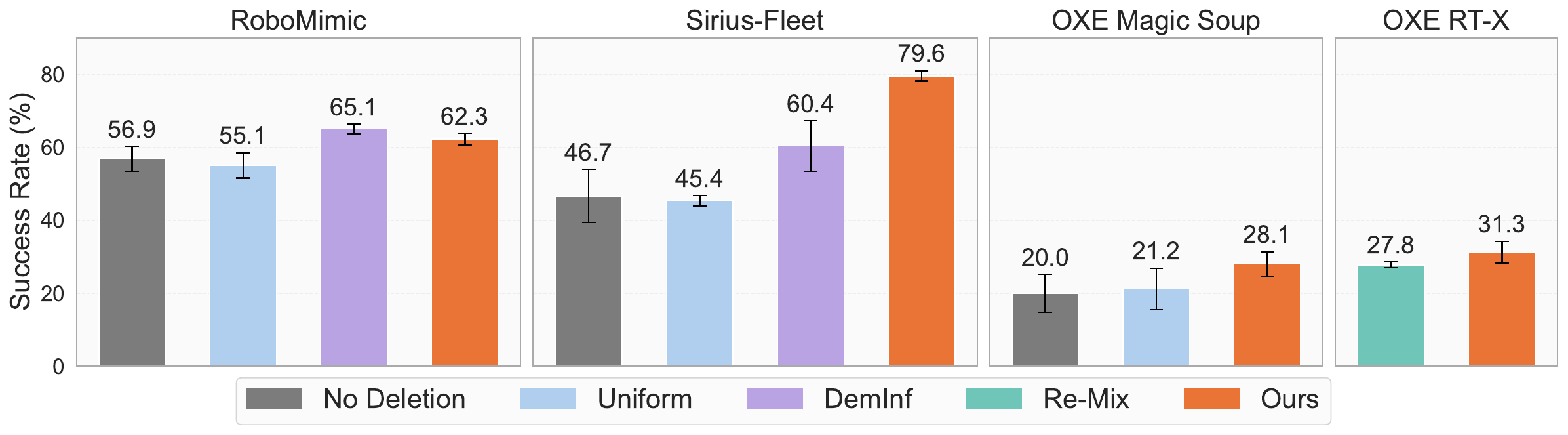}
    \caption{\textbf{Performance comparison across different datasets.} 
    We use the unified threshold for \textsc{Scizor} and report success rates on $4$ datasets. We found that \textsc{Scizor} achieves the strongest performance and outperforms the baselines.
    }
    \label{fig:main_results}
    \vspace{-10pt}
    
\end{figure}
Figure \ref{fig:main_results} summarizes \methodname{}’s impact on policy success rates across all three benchmarks.  Compared to training on the full dataset, \methodname{} delivers absolute gains of $5.4\%$ on RoboMimic, $8.1\%$ on OXE\textsubscript{Magic}, and $32.9\%$ on the Sirius-Fleet real-robot tasks. It also surpasses uniform curation by $16.1\%$ on average, indicating that \methodname{} has a targeted selection of samples to be deleted. These improvements demonstrate that \methodname{}'s data curation consistently filters out low-quality samples and improves policy learning in both simulated and real-world robotic environments.

\textbf{RQ2: What advantage does \methodname{}'s fine-grained state-action curation offer over trajectory- or dataset-level curation in prior work?}
To validate the effectiveness of fine-grained curation on state-action pairs, we compare \methodname{} with two baseline methods: a trajectory-level curation method, Deminf \cite{hejna2025robotdatacurationmutual}, and a dataset-level curation method, Re-Mix \cite{Xie2023DoReMiOD}. Deminf estimates the average contribution of a trajectory towards the mutual information between states and actions in the entire dataset. Re-Mix treats each subset of data as a different “domain” and uses a distributionally robust optimization technique to assign weights to sub-datasets. To ensure a fair comparison, we apply \methodname{} to the same RT-X mixture setting used by Re-Mix. As shown in Figure \ref{fig:main_results} \methodname{} outperforms Re-Mix by $3.5\%$ on average. In the RoboMimic dataset, \methodname{} has not outperformed DemInf, as the dataset is explicitly divided into three levels of trajectory quality, making trajectory-level filtering particularly effective. In contrast, \methodname{} significantly outperforms DemInf by $19.2\%$ on the Sirius-Fleet dataset, where the mixed sources of policy and human actions result in uneven data quality distribution. This suggests that fine-grained state-action curation may be beneficial in datasets with complex and uneven quality distributions.

\vspace{-5pt}
\begin{table}[ht!]
\centering
\caption{\textbf{Ablation studies}: Performance comparison across three datasets (RoboMimic, Sirius-Fleet, and OXE). Our approach consistently outperforms partial ablations, highlighting the importance of combining both components.}
\label{tab:ablation_subop_dedup}
\begin{tabular}{@{}lccc@{}}
\toprule
& \textbf{RoboMimic} & \textbf{Sirius-Fleet} & \textbf{OXE\textsubscript{Magic}}   \\ \midrule
Suboptimal-Removal Only &         60.9 ± 1.8          & 64.2 ± 2.6          & 25.3 ± 2.9  \\
Deduplication Only      &         48.3 ± 0.8         & 63.3 ± 6.9         & 22.1 ± 0.9 \\
\methodname{} (Ours)    & \textbf{62.3 ± 1.6} & \textbf{79.6 ± 1.4} & \textbf{28.1 ± 3.3} \\ \bottomrule
\end{tabular}
\end{table}

\vspace{-8pt}
\textbf{RQ3: What design components contribute most to \methodname{}?}
We first ablate suboptimal data removal and deduplication in Table~\ref{tab:ablation_subop_dedup}. We run experiments only removing suboptimal data or duplicated data, and remove the same amount of data in each dataset as \methodname{}. We find both suboptimal removal and deduplication individually lead to improvements over the baseline, but neither alone is sufficient to match the full performance of \methodname{}. Suboptimal removal is generally more effective than deduplication, but combining both components leads to the largest gains across all datasets.

\vspace{-5pt}
\begin{table}[ht!]
\caption{\textbf{Variations of \methodname{}'s suboptimal data strategies}: We evaluate different scoring strategies for suboptimal data removal: (i) without mixture of transition and trajectory scores, (ii) without temporal discounting, and (iii) the full proposed method (Ours). Results are reported across four tasks, showing that the full version consistently outperforms the alternatives.}
\label{tab:ablation}
\begin{tabular}{@{}lcccc@{}}
\toprule
                  & \textbf{\small RoboMimic Can} & \textbf{\small RoboMimic Square} & \textbf{\small OXE\textsubscript{RT-1} Pick} & \textbf{\small OXE\textsubscript{RT-1} Move} \\ \midrule
\methodname{} w/o mixture  & 81.3 ± 0.6             & 36.0 ± 1.4                & 21.8 ± 7.9            & 12.4 ± 4.6            \\
\methodname{} w/o discount & 79.6 ± 1.4             & 31.5 ± 5.5                & 20.7 ± 6.4            & 9.4 ± 1.4             \\
\methodname{} (Ours)              & \textbf{87.3 ± 0.7}    & \textbf{37.2 ± 2.5}       & \textbf{30.9 ± 8.4}   & \textbf{17.5 ± 1.0}   \\ \bottomrule
\end{tabular}
\end{table}
\vspace{-8pt}
We further investigate \methodname{}'s scoring strategy for suboptimal data classifier in Table 2 by ablating two key components: (i) the transition–trajectory score mixture and (ii) temporal discounting discussed in Section \ref{sec:method-subop}. We train Octo on the OXE ``RT-1" variant \cite{brohan2023rt1roboticstransformerrealworld} with three seeds for faster iteration. Omitting either component consistently degrades performance across all four tasks, highlighting their importance. Temporal discounting lets \methodname{} propagate evidence of suboptimality backward in time, so that transitions leading to poorer future states can also be identified in addition to directly poor actions. The mixture of transition-level and trajectory-level scores balances these fine‐grained penalties with an overall assessment of each demonstration’s quality, making it easier to filter out inherently low‐quality data (for example, trajectories recorded by non‐expert operators). Together, these mechanisms yield the strongest gains in suboptimal data removal.

\begin{wrapfigure}{r}{0.46\textwidth}  
  \vspace{-12pt}                        
  \centering
  \includegraphics[width=\linewidth]{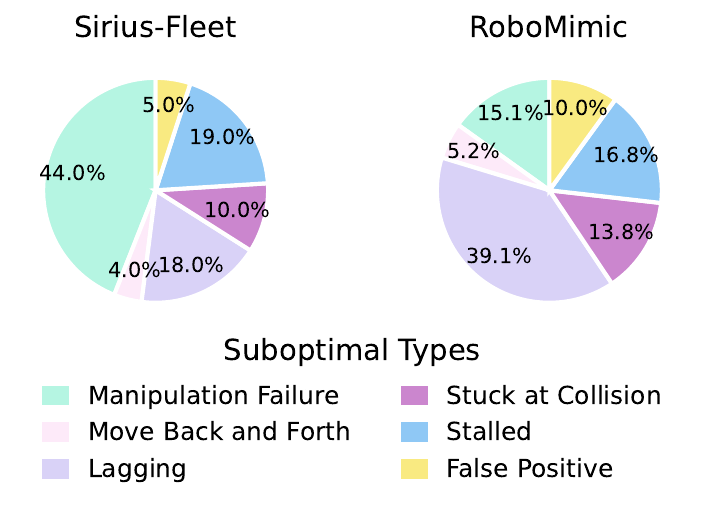}
  \caption{\textbf{Breakdown of suboptimal types classified by \methodname{}.}
  The three dominant failure modes predicted by \methodname{}'s suboptimal classifier are
  \textit{Laggy Motion}, \textit{Manipulation Failure}, and \textit{Pause}, showing \methodname{} removes semantically meaningful transitions.}
  \label{fig:subop_pie_chart}
  \vspace{-5pt}
\end{wrapfigure}
\textbf{RQ4: What types of low-quality samples can \methodname{} identify and curate?}
To qualitatively visualize the suboptimal data identified, we examine the predicted low-quality data and investigate the types of low-quality data they represent. From each of the RoboMimic and Sirius-Fleet datasets, we randomly select $100$ demonstrations flagged with at least one suboptimal segment. We then manually visualize and classify every suboptimal segment across these trials to generate the pie chart. Figure \ref{fig:subop_pie_chart} illustrates the distribution of suboptimal transitions identified by \methodname{}. \textit{Manipulation Failure} refers to errors during grasping---for example, failed grasps, accidental drop of objects. \textit{Pause} denotes transitions with no movement. \textit{Stuck at Collision} describes cases where the gripper or held object collides, leading to a halt. \textit{Lagging} captures motion that proceeds noticeably below the normal speed for the same context. This doesn't contain motions that are intentionally slow, careful, and deliberate.  \textit{Move Back and Forth} indicates aimless motions that don't contribute to task progress. \textit{False Positive} labels misclassified or ambiguous transitions. The most significant fractions are \textit{Laggy}, \textit{Manipulation Failure}, and \textit{Stalled}, indicating that the task-progress classifier identifies semantically meaningful errors rather than spurious noise. Appendix~\ref{sec:exp_and_vis_detail} visualizes representative examples. 

\vspace{-8pt}
\section{Conclusion}
\vspace{-8pt}
We introduce \methodname{}, a self-supervised data curation method that filters suboptimal and redundant state-action pairs to improve imitation learning performance. It combines a task progress predictor to remove suboptimal frames with a similarity-based deduplication module to eliminate overrepresented patterns. By curating the dataset, \methodname{} consistently enhances policy performance across diverse imitation learning benchmarks and outperforms other data curation approaches on large datasets. Future work could explore more adaptive thresholding strategies to achieve optimal deletion ratios and improve the representation of state-action pairs for better curation performance.
\vspace{-8pt}
\section{Limitation}
\vspace{-8pt}


While \methodname{} improves policy success in imitation learning, it has several limitations, which we discuss in detail below:
\vspace{-5pt}


\textbf{Deduplication Representation:} \methodname{}'s deduplication module currently concatenates action and state features. While it performs well in our experiments, future work could explore more expressive or learned representations \cite{pertsch2025fast, li2025unified} that better integrate the action and state spaces.
\vspace{-5pt}


\textbf{Dependence on Demonstration Quality:} \methodname{} assumes that most demonstrations within a trajectory are of good quality, as we rely on self-supervised learning to learn from the majority of the data. If poor-quality demonstrations dominate, the method may become less effective. Future work could focus on better leveraging low-quality data by identifying and utilizing useful segments. 
\vspace{-5pt}



\textbf{Linear Task Progress Assumption:} \methodname{} assumes linear task progression without pausing or repetitive behaviors. However, real-world tasks, like stirring food repeatedly or waiting for it to cook, often involve such behaviors. Future work could adapt the method to better handle these behaviors, e.g., longer history input, hierarchical progress modeling that can capture sub-tasks, multi-timescale progress models that can capture both long-term progress and short-term progress, etc.




\bibliography{main}  

\appendix

\clearpage

\section{Method details}

\label{supp:method_details}

\begin{figure}[h!]
    \centering
    \includegraphics[width=.9\linewidth]{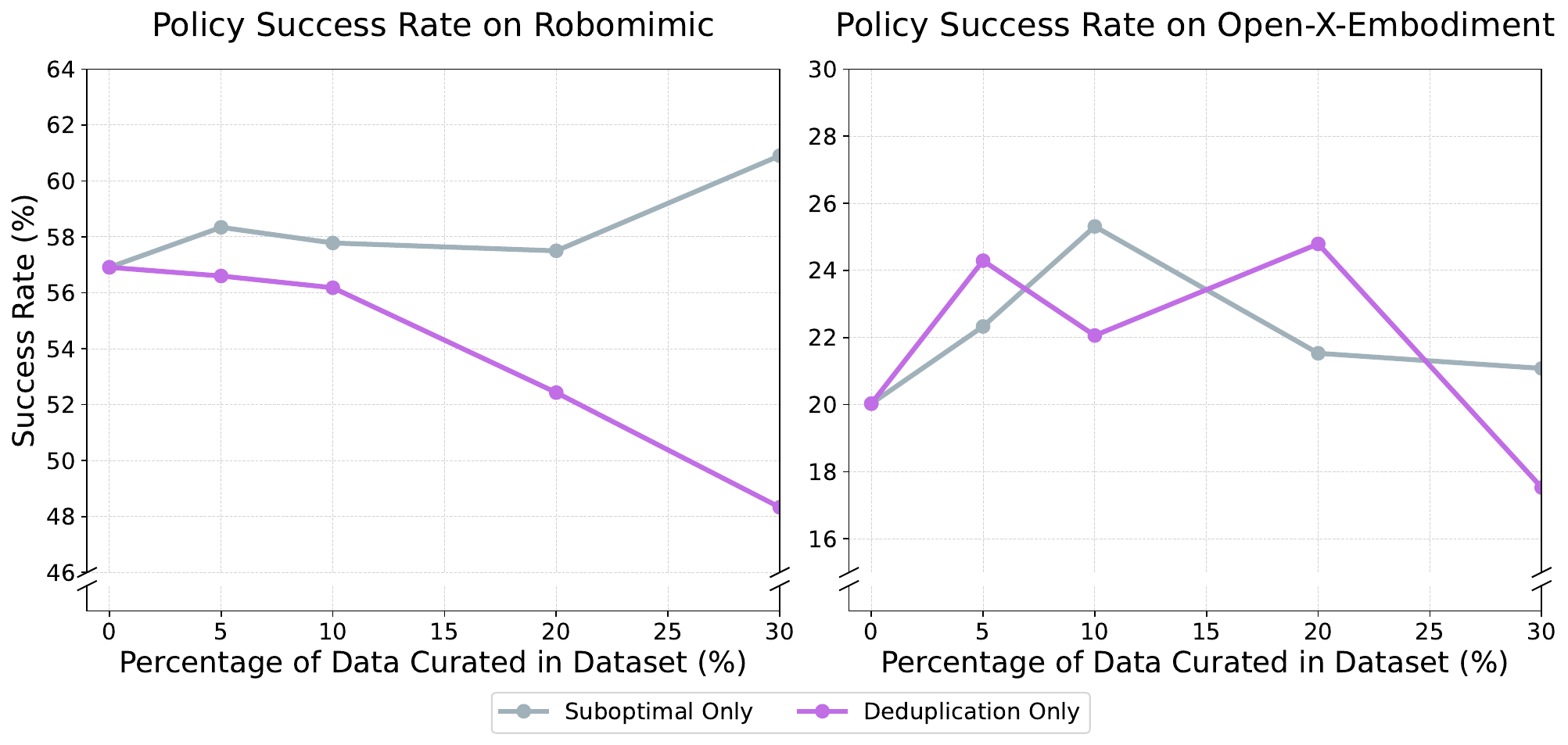}
    \caption{
    The performance of the Suboptimal-Only method and the Deduplication-Only method when deleting different percentages of data on the Robomimic and OXE\textsubscript{Magic} dataset.}
    \label{fig:ratio_vs_success_rate}
\end{figure}

\subsection{Determining the Single Threshold Across Datasets}
 When curating data, we remove all samples with a score above either $\epsilon_{s}$ or $\epsilon_{d}$. To generalize \methodname{} to different datasets, we suggest to find a single threshold. We conduct a hyperparameter search on two simulation datasets, RoboMimic and OXE\textsubscript{Magic} and find that $\epsilon{s} = 0.58$ and $\epsilon_{d} = 0.99$ yield the best performance on both datasets. This single threshold was then directly applied to the real-world setting (Sirius-Fleet). 
 
To find a unified threshold for both suboptimal-transition removal and similarity-based state–action deduplication, we run \methodname{} with only one sub-method at a time on the RoboMimic and OXE\textsubscript{Magic} datasets with deletion ratios of 10\%, 20\%, and 30\%. 

\begin{figure}[h!]
    \centering
    \includegraphics[width=.9\linewidth]{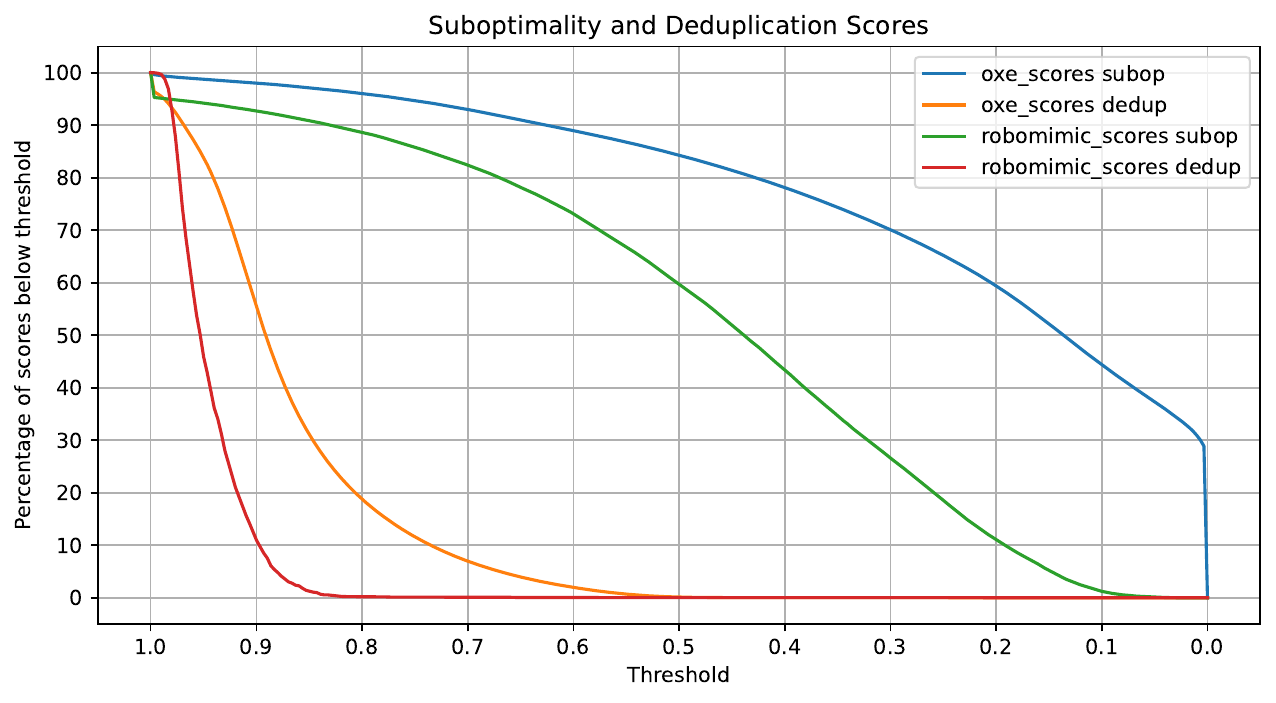}
    \caption{Deletion ratio as a function of the chosen threshold for suboptimal-transition removal and state–action deduplication on the Robomimic and OXE\textsubscript{Magic} datasets.}
    \label{fig:subop_dedup_thres}
\end{figure}

Figure \ref{fig:ratio_vs_success_rate} shows that Suboptimal-Only achieves its highest success rate at 30\% deletion on RoboMimic and 10\% deletion on OXE.
Deduplication-Only performs best with 0\% deletion on RoboMimic and 20\% deletion on OXE.

Next, we plot the deletion ratio as a function of the threshold in Figure \ref{fig:subop_dedup_thres}. We observe that: A suboptimal threshold of 0.58 yields deletion rates of 29.5\% on RoboMimic and 11.9\% on OXE, closely matching their respective optimal ratios.
A deduplication threshold of 0.99 results in only 0.3\% deletion on RoboMimic—insufficient to harm performance—and 4.5\% deletion on OXE, which still provides a notable improvement, very close to deleting $20\%$.

Finally, we apply the unified threshold to all the datasets, and yield the deletion ratio list in Table \ref{tab:deletion_ratio}.

\begin{table}[ht!]
\centering
\caption{\textbf{Deletion Ratio} of all the datasets when unified threshold applied}
\label{tab:deletion_ratio}
\begin{tabular}{@{}lccccc@{}}
\toprule
& \textbf{RoboMimic} & \textbf{Sirius-Fleet} & \textbf{OXE\textsubscript{Magic}} & \textbf{OXE\textsubscript{RT-X}} & \textbf{OXE\textsubscript{RT-1}}   \\ \midrule
Suboptimal-Only &   29.3\%         &   4.7\%       & 11.9\% & 15.0\% & 9.2\% \\
Deduplication-Only      &   0.3\%          &   3.2\%       & 4.5\%  & 0.8\%  & 0.5\% \\
\methodname{} (Total)    & 29.6\%          &   7.9\%       & 15.8\% & 15.8\% & 9.7\% \\ \bottomrule
\end{tabular}
\end{table}

\subsection{Task Progress Prediction Model Architecture.} Given a sub-trajectory $S_{i,i+T}$, the model takes image observation at timesteps $i$ and $i+T$ as inputs. These observations are independently encoded using a frozen DINO-V2 model to obtain the visual features, which provide robust and generalizable visual representations thanks to its self-supervised pretraining on diverse natural images. We compute the difference between the two feature vectors to obtain a delta feature vector, which emphasizes task-relevant changes and accelerates convergence while discarding redundant static information and is concatenated with a CLS token. The feature vector is then processed through a series of multi-layer self-attention transformer blocks. The output CLS token is then fed into a classification head to produce the predicted progress bin.

\subsection{Training the Task Progress Predictor}
We divide each trajectory into five equal time bins. For each training example, we randomly select one bin and sample a time interval \(\Delta t\) uniformly within its bounds. The model takes as input the frame at time \(t\) and the frame at \(t + \Delta t\), and is trained to predict the index of the chosen time bin.

\subsection{Fixed Time Duration}
All experiments use a constant interval of 2s for progress prediction and state–action feature extraction. Since datasets differ in control frequency, this 2s window corresponds to a dataset-dependent number of transitions per second.

\subsection{Task Progress Predicting Details}
Rather than regressing a real-valued progress estimate, we cast progress prediction as classification over discrete temporal bins, which is empirically more robust \citep{aytar2018playing}.
We discretize the temporal gap into $B$ bins, where each bin is a time interval in seconds.
To predict task progress for a sub-trajectory $S_{i,i+T}$, we train a task progress classifier to classify the bin corresponding to the time $T$ between the start and end states.
Empirically, we set $B=5$, sample each sub-trajectory as $2$ seconds, and bins to be $[0, 0.5), [0.5, 1.0), [1.0, 2.0), [2.0, 5.0)$, and $ [5.0, +\infty)$. 
For the weighted combination coefficient between local suboptimality score and the mean score across all time steps, we simply choose 0.5.

\subsection{Removing Suboptimal Samples Details}
Note that each sample preserves an observation history and an action sequence for algorithms that are history-dependent (e.g., BC-Transformer) and that utilize action chunking. We do the chunking before any removal to avoid temporal inconsistencies.

\subsection{Deduplication Details}
We use a chunk size of N=8 for deduplication, as it's commonly supported by most of the video encoder.

\section{Experimental Details}

\subsection{Dataset Details \label{sec:dataset_detail}}
\textbf{RoboMimic \citep{mandlekar2021matterslearningofflinehuman}} is a robotic imitation learning dataset and benchmark. It provides trajectories collected by proficient human (PH) or mixed-proficient human (MH) demonstrators. The PH dataset consists of 200 trajectories from a single experienced demonstrator, while the MH dataset includes 300 trajectories from six demonstrators --- two ``better", two ``okay", and two ``worse". For our experiment, we use the MH dataset for the ``Can" and ``Square" tasks.

\textbf{Sirius-Fleet \citep{liu2024multitaskinteractiverobotfleet}} uses a visual world model to predict sub-optimal behaviors during policy rollout and requests human intervention when needed. The \textbf{Sirius-Fleet} dataset is collected over three rounds by allowing the policy to roll out and incorporating human-corrected data for retraining. We utilize the real-world \textbf{Sirius-Fleet} dataset, which adopts the Mutex settings \citep{shah2023mutex} and includes 1,500 trajectories. Our real-world evaluation spans four task sets comprising eight tasks.

\textbf{Open-X-Embodiment (OXE) \citep{padalkar2023open}} is a large-scale collection of over one million real-world robotic trajectories. The dataset is multi-task and cross-embodiment, covering various action and observation spaces. We use the RLDS Dataset Modification \citep{rlds_dataset_mod} to unify the action space to 7 DoFs. We employ three variations of the OXE dataset, each selecting different subsets of the original data and applying different weightings: the ``Magic Soup" mixture (OXE\textsubscript{Magic}) used in the Octo paper \citep{octomodelteam2024octoopensourcegeneralistrobot}, the ``RT-X" mixture (OXE\textsubscript{RT-X}) used in the Re-Mix paper \citep{Xie2023DoReMiOD}, and the ``RT-1" dataset (OXE\textsubscript{RT-1}) from the RT-1 paper \citep{brohan2023rt1roboticstransformerrealworld}.

\subsection{Training and Evaluation Details}  

\textbf{Policy Training Details}
\label{sec:train_and_eval_detail}

\begin{table}[ht!]
\centering
\caption{Hyperparameter configurations and architectural details for the Robomimic, Sirius-fleet, and OXE datasets used in our experiments.}
\label{tab:hyperparameter}
\resizebox{\linewidth}{!}{
\begin{tabular}{lccccc}
\toprule
\textbf{} & \textbf{RoboMimic} & \textbf{Sirius-Fleet Real} & \textbf{OXE\textsubscript{Magic}} & \textbf{OXE\textsubscript{RT-X}} & \textbf{OXE\textsubscript{RT-1}} \\
\midrule
Architecture    & BC                 & BC-Transformer-GMM & Octo & Octo & Octo \\
Learning Rate   & 1e-4               & 1e-4               & 3e-4 & 3e-4 & 3e-4 \\
Weight Decay    & 0.1                & 0.1                & 0.1  & 0.1  & 0.1  \\
Batch Size      & 16                 & 16                 & 2048 & 2048 & 256  \\
Params          & 23M                & 35M                & 93M  & 93M  & 93M  \\
Steps           & 300K               & 1M                 & 300K & 300K & 200K \\
Action Chunk    & 1                  & 10                 & 4    & 4    & 4    \\
Obs History     & 1                  & 10                 & 2    & 2    & 2    \\
GPU             & 1 L40S 48GB        & 1 L40S 48GB        & 32 H100 80GB & 32 H100 80GB & 8 L40 48GB \\
\bottomrule
\end{tabular}
}
\end{table}

We evaluate \methodname{} across various architectures and datasets to demonstrate its applicability to different imitation learning algorithms. We intentionally leave all model architectures and hyper-parameters unchanged from the public reference implementations of each dataset, demonstrating that \methodname{} is \emph{plug-and-play}.  For the \textbf{RoboMimic} and \textbf{Sirius-Fleet} experiments, we train each model using three random seeds. For the larger \textbf{OXE} experiments, we train the Octo model using two seeds.

For each setting, the same dataset is used throughout SCIZOR suboptimal classifier training, deduplication, and subsequent policy training. For example, in the OXE setting, we first train SCIZOR’s task progress predictor on the full OXE dataset and use it to identify suboptimal transitions within that dataset. After filtering out low-quality segments, we apply de-duplication to the remaining data. The final policy is then trained on this curated dataset. Our pipeline is entirely self-supervised, requiring no explicit data quality labels for training.

On the \textbf{RoboMimic} dataset, we train a basic Behavior Cloning (BC) model with MLP layers \citep{mandlekar2021matterslearningofflinehuman} for 600 epochs. We evaluate every 20 epochs, select the top three checkpoints per random seed, and report the mean success rate over 80 trials per task for each checkpoint, then average across seeds.

For the \textbf{Sirius-Fleet} real-robot experiments, we use a BC-Transformer model with a GMM head \citep{mandlekar2021matterslearningofflinehuman, liu2022robot, liu2024multitaskinteractiverobotfleet}, and train for 2000 epochs. Evaluation is performed at the 2000 epoch, and we report the average success rate of the top three checkpoints for each seed. We run 10 trials per task per seed for this setting.

We train the Octo-Small model on all three variations of the \textbf{OXE} dataset. For \textbf{OXE\textsubscript{Magic}} and \textbf{OXE\textsubscript{RT-X}}, training is performed for 300K steps with a batch size of 2048 using two seeds. For \textbf{OXE\textsubscript{RT-1}}, we train for 200K steps with a batch size of 256 using three seeds. Evaluation is conducted in the SIMPLER simulation environment \citep{li24simpler} on the ``Pick Coke Can" and ``Move Near" tasks. We only evaluate on the Visual-Matching setting of SIMPLER, which means there won't be lighting or texture variation. For each seed, we identify the highest success rate among the last three saved checkpoints, and then average these best performances across seeds. We evaluate 300 trials per checkpoint on ``Pick Coke Can" and 240 trials per checkpoint on ``Move Near".

Table \ref{tab:hyperparameter} provides the detailed hyperparameter and model architectures used in our experiments.

\subsection{Other Simpler Baselines}

We further compare \methodname{} against two simple baselines. To show the necessity of progress estimation, we design a velocity filter: we compute the end-effector velocity as the L2 norm of the delta end effector state and apply a moving-window filter to remove the slowest 
k\% of transitions. To evaluate whether zero-shot VLMs can serve as effective demonstration raters, we prompt GPT-4.1-small to categorize each demonstration into three ranks and discard the lowest
k\%. Results in Table~\ref{tab:plain_baselines} show that only \methodname{} improves over the No Curation baseline. The moving window
velocity filter catches only slow movements; once those are
deleted, further pruning harms performance. The VLM baseline
helps on the “Can” task but hurts on “Square”, indicating its
limited spatial understanding for precise, fine-grained assembly.


\begin{table}[ht]
\centering
\begin{tabular}{lcccc}
\toprule
 & Velocity Filter & Zero-Shot VLM & No Curation & \textbf{\methodname{}} \\
\midrule
Can (\%)    & $75.7 \pm 1.3$ & $80.4 \pm 4.0$ & $79.0 \pm 1.7$ & $\mathbf{87.3 \pm 0.7}$ \\
Square (\%) & $30.1 \pm 3.2$ & $32.1 \pm 2.4$ & $34.8 \pm 5.1$ & $\mathbf{37.2 \pm 2.5}$ \\
\bottomrule
\end{tabular}
\caption{Comparison of methods on Robomimic.}
\vspace{-3mm}
\label{tab:plain_baselines}
\end{table}

\subsection{Additional Research Questions}

\textbf{RQ5: What are the false negatives and false positives?}
We analyze the suboptimal cases that were incorrectly classified by \methodname{}, labeled as false positive in Figure~\ref{fig:subop_pie_chart}. Among the $5\%$ of false positive segments predicted in Sirius-Fleet, $3\%$ occur near the end of the demonstrations. These segments often include repeated padded frames, since the task has already been completed. The remaining 2\% are short segments right before a suboptimal pause, which would also be predicted as suboptimal segments.
To conduct a controlled error analysis, we created a human-annotated dataset of robot failures from the Sirius-Fleet dataset. 13/25 of failures are successfully predicted. The false negatives can be categorized into two types: (1) The robot failed to grasp at first, but it recovered quickly and didn't cause a big progress lag. This is because our curation threshold focuses on deleting the most detrimental suboptimality. (2) The self-supervised model failed to detect because most of the data contains this failure.

\textbf{RQ6: How well can \methodname{} preserve high quality data?}
We investigate whether SCIZOR removes high-quality data during the curation process through the following analyses. 1) Keeping Expert Demonstrations. We randomly picked 50 perfect expert demonstrations from 5 tasks from Sirius-Fleet, each containing 10 demonstrations. Surprisingly,none of these demonstrations have SCIZOR-predicted suboptimal segments, indicating that SCIZOR is actually considering them as high-quality samples and won’t be curated. 2) Keeping Recovery Behaviors. We also checked the grasping failures mentioned in the previous paragraph. Specifically, 54\% of recovery behaviors are fully maintained, and 31\% are partially maintained. Only 15\% are deleted, and the remaining episodes directly start with the next rasping behavior. Most of the recovering behaviors are kept because they contribute to task progress.

\subsection{Evaluation Task Details for each Dataset}
Table \ref{tab:task_details} presents the detailed descriptions and visualizations of the tasks used in our experimental settings. These tasks span both simulated and real-world environments, covering a diverse range of manipulation challenges.

\begin{center}
\begin{longtable}{@{} 
    >{\rule{0pt}{4\baselineskip}\centering\arraybackslash}m{0.19\textwidth}   
    >{\rule{0pt}{4\baselineskip}}p{0.75\textwidth}                             
  @{}}
  \caption{Task Name and Description for each setting.}
  \label{tab:task_details}\\

  \toprule
  \multicolumn{1}{p{0.19\textwidth}}{\centering\textbf{Visualization}}
    & \multicolumn{1}{p{0.75\textwidth}}{\centering\textbf{Task Name and Description}} \\
  \midrule
  \endfirsthead

  \multicolumn{2}{@{}l}{\small\itshape continued from previous page}\\
  \toprule
  \multicolumn{1}{p{0.19\textwidth}}{\centering\textbf{Visualization}}
    & \multicolumn{1}{p{0.75\textwidth}}{\centering\textbf{Task Name and Description}} \\
  \midrule
  \endhead

  \midrule
  \multicolumn{2}{r}{\small\itshape continued on next page} \\
  \endfoot

  \bottomrule
  \endlastfoot

  \includegraphics[width=\linewidth]{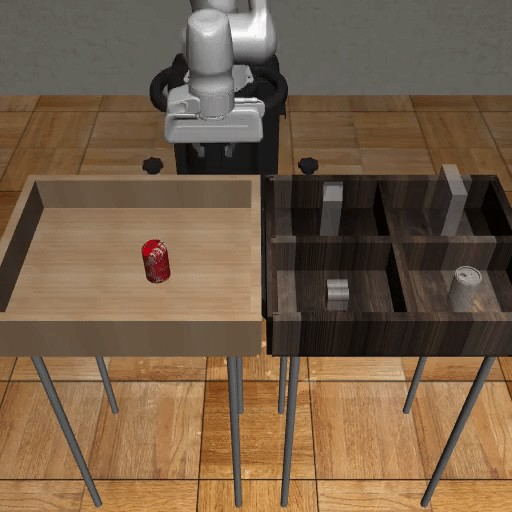}
    & \parbox[c][4\baselineskip][c]{\linewidth}{\textbf{Robomimic Can}\\
        The robot is required to place a coke can from a large bin into a smaller target bin.}\\
  \midrule

  \includegraphics[width=\linewidth]{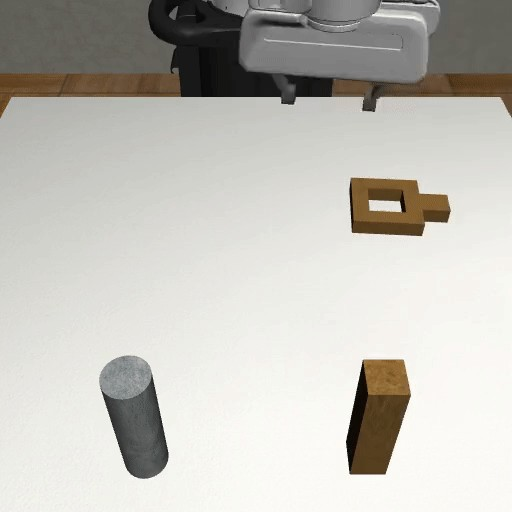}
    & \parbox[c][4\baselineskip][c]{\linewidth}{%
        \textbf{Robomimic Square}\\
        The robot is required to pick a square nut and place it on a rod.
        Substantially more difficult than Pick Place Can due to the precision required.
      } \\
  \midrule

  \includegraphics[width=\linewidth]{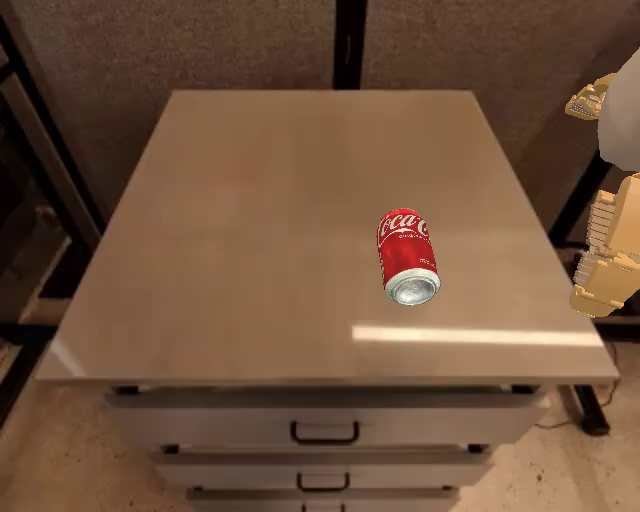}
    & \parbox[c][4\baselineskip][c]{\linewidth}{%
        \textbf{SIMPLER Pick Coke Can}\\
        The robot is instructed to grasp the empty coke can on the table and lift it up.
      } \\
  \midrule

  \includegraphics[width=\linewidth]{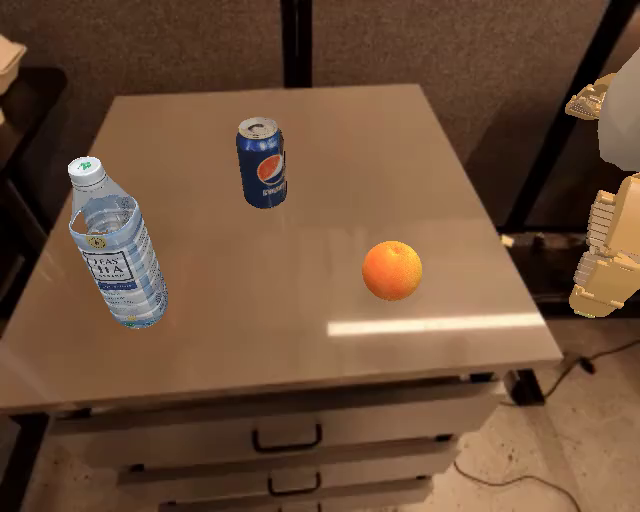}
    & \parbox[c][4\baselineskip][c]{\linewidth}{%
        \textbf{SIMPLER Move Near}\\
        The robot is instructed to move one object next to another object, while the third object serves as a distractor.
      } \\
  \midrule

  \includegraphics[width=\linewidth]{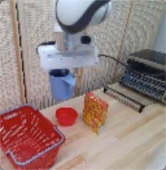}
    & \parbox[c][4\baselineskip][c]{\linewidth}{%
        \textbf{Mutex Mug to Basket}\\
        The robot is instructed to pick up the blue mug and then place it in the basket.
      } \\
  \midrule

  \includegraphics[width=\linewidth]{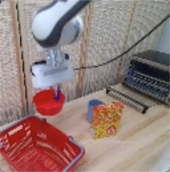}
    & \parbox[c][4\baselineskip][c]{\linewidth}{%
        \textbf{Mutex Bowl to Basket}\\
        The robot is instructed to pick up the red bowl and then place it in the basket.
      } \\
  \midrule

  \includegraphics[width=\linewidth]{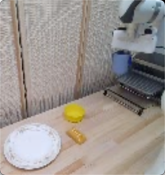}
    & \parbox[c][4\baselineskip][c]{\linewidth}{%
        \textbf{Mutex Mug to Oven Tray}\\
        The robot is instructed to pick up the blue mug and then place it on the oven tray.
      } \\
  \midrule

  \includegraphics[width=\linewidth]{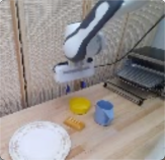}
    & \parbox[c][4\baselineskip][c]{\linewidth}{%
        \textbf{Mutex Bread to Plate}\\
        The robot is instructed to pick up the bread and then place it on the white plate.
      } \\
  \midrule

  \includegraphics[width=\linewidth]{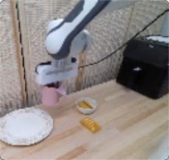}
    & \parbox[c][4\baselineskip][c]{\linewidth}{%
        \textbf{Mutex Mug to Plate}\\
        The robot is instructed to pick up the pink mug and then place it on the white plate.
      } \\
  \midrule

  \includegraphics[width=\linewidth]{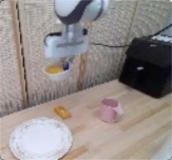}
    & \parbox[c][4\baselineskip][c]{\linewidth}{%
        \textbf{Mutex Bowl to Plate}\\
        The robot is instructed to pick up the bowl with hot dogs and then place it on the white plate.
      } \\
  \midrule

  \includegraphics[width=\linewidth]{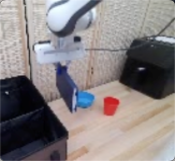}
    & \parbox[c][4\baselineskip][c]{\linewidth}{%
        \textbf{Mutex Book to Caddy}\\
        The robot is instructed to pick up the book and then place it in the back compartment of the caddy.
      } \\
  \midrule

  \includegraphics[width=\linewidth]{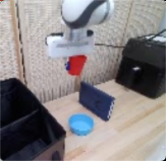}
    & \parbox[c][4\baselineskip][c]{\linewidth}{%
        \textbf{Mutex Cup to Caddy}\\
        The robot is instructed to pick up the red cup and then place it in the front compartment of the caddy.
      } \\

\end{longtable}
\end{center}

\subsection{Extra results and Visualization}
\label{sec:exp_and_vis_detail}
Table \ref{tab:full_results_robomimic}, \ref{tab:sirius_full_results}, \ref{tab:full_results_oxe} presents the detailed results of different methods on each task of the reported datasets.
\begin{table}[htbp]
\centering
\caption{Success rates on Robomimic across different tasks and methods}
\label{tab:full_results_robomimic}
\begin{tabular}{lcc}
\toprule
& \textbf{Can} & \textbf{Square} \\
\midrule
Suboptimal-Removal Only & 84.0\%            & 37.8\%          \\
Deduplication Only      & 74.3\%            & 22.4\%          \\
Random Deletion         & 78.0\%            & 32.2\%          \\
DemInf                  & \textbf{88.9}\%   & \textbf{41.4}\% \\
\methodname{} (Ours)    & 84.0\%            & \textbf{40.8}\% \\
\bottomrule
\end{tabular}
\end{table}

\begin{table}[htbp]
\centering
\caption{Success rates on Sirius-Fleet across different tasks and methods}
\label{tab:sirius_full_results}
\resizebox{\linewidth}{!}{
\begin{tabular}{lcccc}
\toprule
 & Book→Caddy & Cup→Caddy & Bowl→Plate & Mug→Plate \\
\midrule
No Deletion                  & 40.0\% & 65.0\% & 65.0\% & 35.0\% \\
Suboptimal-Removal Only      & 65.0\% & \textbf{75.0\%} & 70.0\% & 60.0\% \\
Deduplication Only           & 45.0\% & \textbf{75.0\%} & 80.0\% & 70.0\% \\
Random Deletion              & 50.0\% & 65.0\% & 55.0\% & 30.0\% \\
Deminf                       & \textbf{66.7\%} & 53.3\% & 60.0\% & 50.0\% \\
\methodname{} (Ours)         & \textbf{66.7\%} & 73.3\% & \textbf{93.3\%} & \textbf{83.3\%} \\
\midrule
 & Mug→Basket & Bowl→Basket & Mug→Tray & Bread→Plate \\
\cmidrule(lr){1-5}
No Deletion                  & 35.0\% & 60.0\% & 35.0\% & 50.0\% \\
Suboptimal-Removal Only      & 35.0\% & 90.0\% & 50.0\% & 60.0\% \\
Deduplication Only           & 65.0\% & 70.0\% & 60.0\% & 60.0\% \\
Random Deletion              & 25.0\% & 70.0\% & 30.0\% & 35.0\% \\
Deminf                       & 50.0\% & 76.7\% & 53.3\% & 63.3\% \\
\methodname{} (Ours)         & \textbf{66.7\%} & \textbf{96.7\%} & \textbf{66.7\%} & \textbf{90.0\%} \\
\bottomrule
\end{tabular}
}
\end{table}

\begin{table}[htbp]
\centering
\caption{Success rates on OXE across different tasks, mixtures and methods}
\label{tab:full_results_oxe}
\begin{tabular}{lcc}
\toprule
       & \textbf{Pick Can}  & \textbf{Move Near} \\
\midrule
\multicolumn{3}{c}{OXE\textsubscript{Magic} Mixture} 
\\
No Deletion            & 27.0\%       & 13.1\%       \\
Random Deletion                 & 29.1\%       & 12.1\%       \\
Suboptimal-Removal Only                  & 33.7\%       & \textbf{16.9}\%       \\
Deduplication Only                  & 31.8\%       & 12.3\%       \\
\methodname{} (Ours)                   & \textbf{39.5}\%       & \textbf{16.7}\%       \\
\midrule
\multicolumn{3}{c}{OXE\textsubscript{RT-X} Mixture}     \\
Remix              & 40.5\%       & 15.0\%       \\
\methodname{}(Ours)& \textbf{43.3}\%       & \textbf{19.2}\%       \\
\bottomrule
\end{tabular}
\end{table}

\begin{figure}[h!]
    \centering
\includegraphics[width=.9\linewidth]{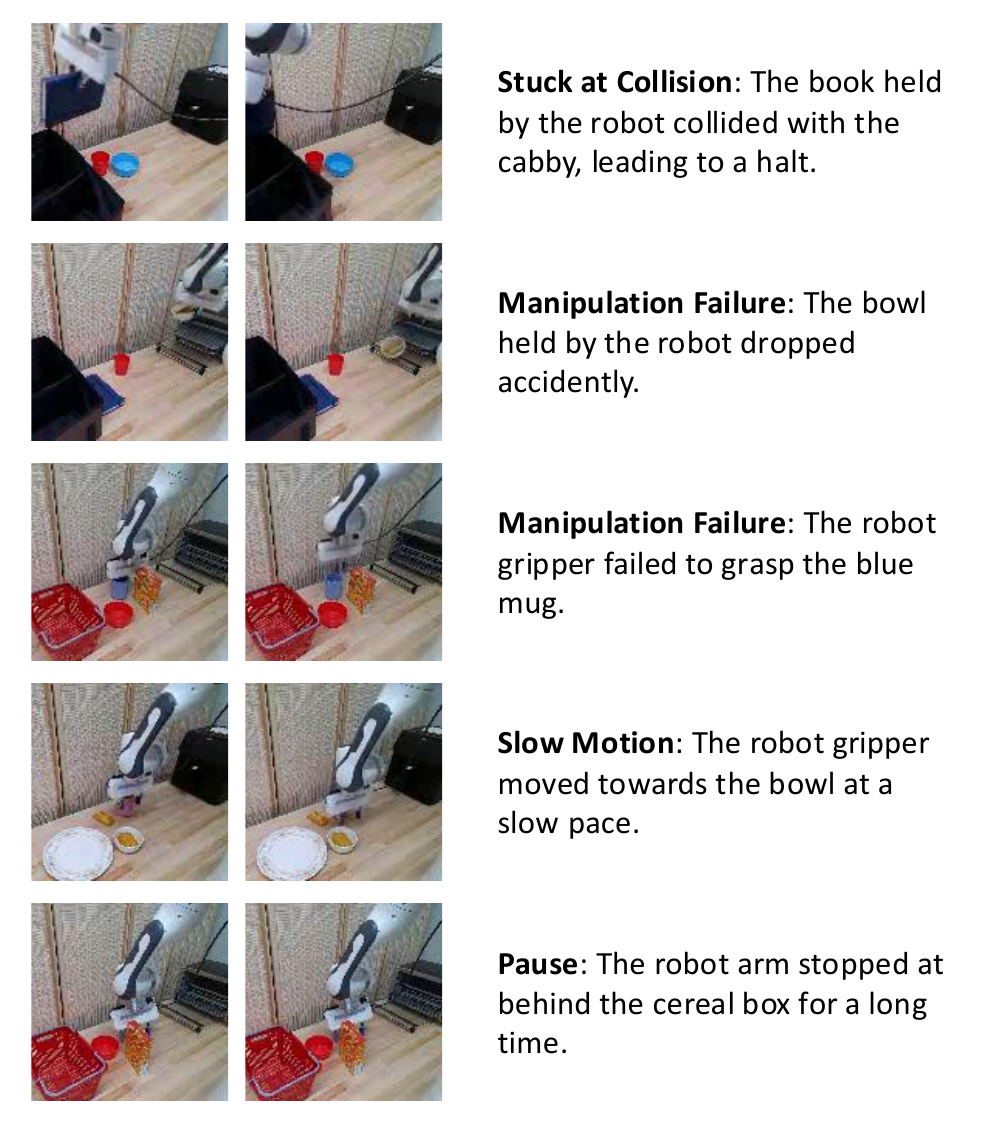}
    \caption{
    Visualizations for suboptimal scenarios detected in Robomimic dataset}
    \label{fig:robomimic_subop_vis}
\end{figure}

\begin{figure}[h!]
    \centering
    \includegraphics[width=.9\linewidth]{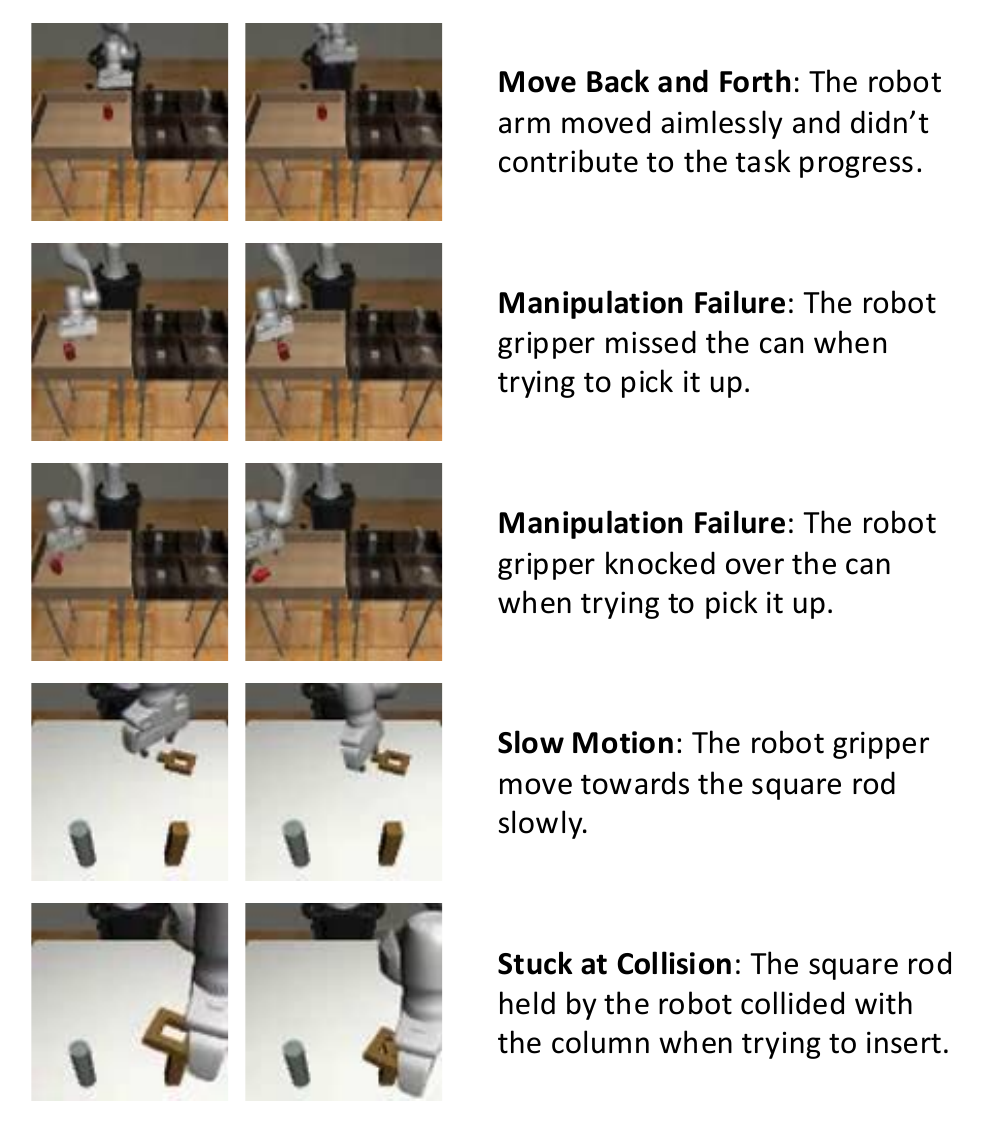}
    \caption{
    Visualizations for suboptimal scenarios detected in Sirius-Fleet dataset}
    \label{fig:mutex_subop_vis}
\end{figure}

We also visualize the suboptimal scenarios identified by \methodname{} across both the Robomimic and Sirius-Fleet datasets in Figure \ref{fig:robomimic_subop_vis}, \ref{fig:mutex_subop_vis}. These visualizations highlight common suboptimal modes detected by our method, offering insight into \methodname{}’s capabilities.

\subsection{Findings during Evaluation}
During evaluation of \methodname{} on the Sirius-Fleet dataset, we observed that several failure modes present in the policy trained on the full dataset disappeared when using \methodname{} to curate the dataset.

For example, in the book-to-caddy task, the baseline policy often allowed the book to collide with the caddy, whereas our policy reliably avoids any contact. Furthermore, our policy is noticeably more reactive: when it does encounter a failure, the baseline tends to pause or oscillate in place, but our policy quickly returns to its original trajectory and retries until the task is completed.

\end{document}